\documentclass[10pt,twocolumn,letterpaper]{article} 

\usepackage{avss}
\usepackage{times}
\usepackage{epsfig}
\usepackage{graphicx}
\usepackage{amsmath}
\usepackage{multirow}
\usepackage{amssymb}
\usepackage{caption}
\usepackage{subcaption}
\usepackage{booktabs}

\usepackage{todonotes}

\usepackage[pagebackref=true,breaklinks=true,letterpaper=true,colorlinks,bookmarks=false]{hyperref}

\avssfinalcopy 

\def\avssPaperID{86} 

\ifavssfinal\fi
\begin{document}

\title{Detection-aware multi-object tracking evaluation}
\author{Juan C. SanMiguel, Jorge Muñoz \\
Video Processing and Understanding Lab, Universidad Aut\'{o}noma de Madrid, Madrid (Spain)\\
{\tt\small \{juancarlos.sanmiguel@uam.es, jorge.munnoza@estudiante.uam.es\} \thanks{This work is part of the preliminary tasks related to the Harvesting Visual Data (HVD) project (PID2021-125051OB-I00) funded by the Ministerio de Ciencia e Innovación of the Spanish Government.}}
\and
Fabio Poiesi\\
Technologies of Vision, Fondazione Bruno Kessler, Trento (Italy)\\
{\tt\small \{poiesi@fbk.eu\}}
\thanks{This work was also supported by the H2020 Framework Programme through the project MiMEx under Grant 965486.}
}

\maketitle
\thispagestyle{empty}  


\begin{abstract}
How would you fairly evaluate two multi-object tracking algorithms (i.e.~trackers), each one employing a different object detector? 
Detectors keep improving, thus trackers can make less effort to estimate object states over time. 
Is it then fair to compare a new tracker employing a new detector with another tracker using an old detector? 
In this paper, we propose a novel performance measure, named Tracking Effort Measure (TEM), to evaluate trackers that use different detectors. 
TEM estimates the improvement that the tracker does with respect to its input data (i.e.~detections) at frame level (intra-frame complexity) and sequence level (inter-frame complexity).
We evaluate TEM over well-known datasets, four trackers and eight detection sets. 
Results show that, unlike conventional tracking evaluation measures, TEM can quantify the effort done by the tracker with a reduced correlation on the input detections.
Its implementation will be made publicly available online.\footnote{\url{https://github.com/vpulab/MOT-evaluation}}.
\end{abstract}
\let\thefootnote\relax\footnote{\textcopyright2022 IEEE. Personal use of this material is permitted. Permission from IEEE must be
obtained for all other uses, in any current or future media, including
reprinting/republishing this material for advertising or promotional purposes, creating new
collective works, for resale or redistribution to servers or lists, or reuse of any copyrighted
component of this work in other works. DOI: 10.1109/AVSS56176.2022.9959412.}

\section{Introduction}
Multiple Object Tracking (MOT) is nowadays a hot research topic for many domains such as video-surveillance and traffic video monitoring. 
Several recent MOT algorithms (trackers) follow the tracking-by-detection paradigm~\cite{chen2022survey,marvasti2021deep}, which considers detection and tracking separately.
While detection aims to locate objects of interest in each frame (e.g.~via bounding boxes), tracking determines the correspondences among detections in all frames of the video sequence with the objective of assigning unique identifiers to these detections (IDs).

Performance evaluation measures and benchmarks are pivotal to evaluate the quality of such trackers~\cite{chen2022survey}
Many measures have been proposed for MOT such as Multiple Object Tracking Accuracy (MOTA)~\cite{kasturi2008framework}, Multiple Extended-target Tracking Error (METE)~\cite{Nawaz2014} and Higher Order Tracking Accuracy (HOTA)~\cite{luiten2021hota}. The situation for detection algorithms is similar to the tracking case. A plethora of detectors exist~\cite{zaidi2022survey}, which are often evaluated using measures such as Precision and Recall~\cite{nghiem2007etiseo}, and mean Average Precision (mAP) \cite{Everingham2010}.

A critical issue arises when existing performance measures are used to compare new and old trackers.
These trackers often use state-of-the-art detectors with increasingly better performance, and therefore, it becomes difficult to estimate the actual improvement of a tracker.
It is unclear whether such improvement comes from a better detector or a better tracker. In this situation, the MOT benchmark~\cite{dendorfer2021motchallenge} provides sets of detections that must be used for all participants (i.e.~\textit{public detections}), so the detector dependency is removed. However, detectors keep improving over time and the performance of these public detections does not represent the state-of-the-art anymore~\cite{padilla2020survey}. As a result, rankings with more recent detectors (i.e. \textit{private detections}) are allowed in the MOT benchmark, exhibiting higher performance as compared to using public detections. Therefore, it would desirable to be able to compare different detector-tracker combinations, as current MOT performance measures are unable to address such comparison fairly \cite{valmadre2021local,luiten2021hota}.

In this paper, we contribute to the abovementioned limitations for performance evaluation of tracking-by-detection algorithms. 
We propose a novel evaluation measure that can estimate and reduce the dependency between detection and tracking. 
In particular, we define the measure \textit{intra-frame complexity} to estimate the improvement that the tracker does over the detector at frame-level. 
We also define the measure \textit{inter-frame complexity} that extends the previous proposal to multiple frames. 
Finally, we combine both proposed measures to get \textit{a final complexity score} to compare different detector-tracker combinations. We validate the proposed measures using two MOT benchmark datasets (MOT17 and MOT20) and 32 detector-tracker combinations (four trackers and eight detection sets). Our findings show that our proposal is independent of detector's performance unlike related MOT performance measures.

This paper is organized as follows: Section \ref{sec:related-work} discusses the related work, in Section \ref{sec:proposal} we overview the proposed performance measures, experimental results are given in Section \ref{sec:evaluation}, and Section \ref{sec:conclusions} concludes the paper.

\section{Related work}\label{sec:related-work}


\subsection{Object detection performance}
The goal of object detection performance measures is to score the detections based the overlap between estimated and ground-truth bounding boxes. 
Conventional measures are True Positives (TP), False Positives (FP) and False negatives (FN) that account for detections matched with ground-truth, detections not matched with ground-truth and missing detections for existing ground-truth \cite{kasturi2008framework}, respectively. 
Matching is typically computed through the Intersection-Over-Union (IOU)~\cite{manohar2006performance}, which quantifies the spatial overlap of the corresponding bounding boxes.
Precision and Recall measure the number of correct detections with respect to the total number of detections and the number of correct detections with respect to the number of ground-truth ones, respectively~\cite{nghiem2007etiseo,conte2010performance}. 
To account for different detector settings leading to different Precision-Recall results, a popular evaluation measure is the Average
Precision (AP) \cite{padilla2020survey}. 


\subsection{Visual tracking performance}
The goal of multi-object tracking performance measures is to quantify the similarity between estimated and ground-truth trajectories, where each trajectory is represented by a unique object identifier (ID). 
Several studies have been proposed to analyse similarities among existing measures~\cite{vcehovin2016visual,fang2017performance}. 
Here, we review some of the most popular tracking performance measures.
 
A typical tracking challenge is the discrimination of objects with similar appearance that may result in identity switches.
The measure IDSW counts the number of identity switches that occur for each estimated trajectory~\cite{ristani2016performance}. 
The Multiple Object Tracking Accuracy (MOTA) combines IDS, FN and FP into a single score~\cite{bernardin2008evaluating}. 
However, MOTA has a few limitations~\cite{valmadre2021local, luiten2021hota, Nawaz2014}: it is dependent on the video frame rate; it is unbounded and can be negative; and its expected behavior is not symmetric in terms of Recall and Precision. 
The IDF1 measure operates at sequence-level by combining Precision and Recall of trajectories~\cite{ristani2016performance}. 
As main limitation of this measure, it requires a localization threshold to determine the matching of trajectories. 
Similarly to IDF1, the Average Tracking Accuracy (ATA) associates estimated and ground-truth bounding boxes at frame and sequence levels.
The Higher Order Tracking Accuracy (HOTA) addresses some limitations of MOTA by performing an evaluation for a set of thresholds, similarly to mAP with Recall and Precision~\cite{luiten2021hota}.
HOTA evaluates with a single score long-term associations and can also be decomposed into sub-measures to facilitate the inspection of detection and tracking (association) tasks. 
The METE measure \cite{Nawaz2014} provides an holistic approach to measure the tracking error that accounts for state of the object, without applying thresholds, and for the cardinality. 
An alternative approach \cite{carvalho2012filling} combines measures based on different types of ground-truth information with the objective of approximating the ideal error. 
Its main limitation is the need for multiple annotations for the same data.
 
From another perspective, the NLL measure accounts for the uncertainty of the tracker output \cite{pinto2021uncertainty}. 
Albeit effective, it requires a specific probabilistic output (multi-object posterior) that may not be generated by many current trackers.
The TW-TM measure operates on sets of trajectories and includes weighting factors to penalize tracking errors (localization error, FP, FN and IDSWs)~\cite{garcia2021time}. 
However, configuring such penalization is not straightforward for any tracker. 
Time-scope restrictions have also been proposed recently for many popular measures \cite{valmadre2021local}, allowing to define the relative importance of detection and association. 
Another measure is proposed in \cite{marcenaro2012performance} which considers three elements at trajectory level: fragmentation, coverage and cardinality. This measure successfully evaluates a specific type of tracker but it is not contrasted against related measures. 
The average time between failures is also employed to evaluate tracking performance \cite{Carr2016}.

Other tracking performance measures focus on extracting quality indicators during run-time execution of the tracker without requiring the full trajectory related to multiple-hypothesis outputs \cite{sanmiguel2012adaptive} or motion-association costs \cite{denman2009dynamic}. 
However, these measures do not employ ground-truth and therefore, their accuracy is lower compared to the previously discussed measures.

In summary, the relationship between detection and tracking in recent trackers has not been properly considered in the related work (with the exception of HOTA) and therefore, existing measures cannot fairly estimate performance for different detector-tracker combinations.
Our measure aims at addressing these limitations by quantifying the quality of multi-object trackers irrespective of the detectors.

\section{Tracking Effort Measure}\label{sec:proposal}

Our Tracking Effort Measure (TEM) aims to estimate the performance of the tracking-by-detection components: detection and tracking. 
We assume that detections are provided in the form of bounding boxes.
We measure the performance at both frame-level (intra-frame complexity) and sequence-level (inter-frame complexity). 
Both measures are combined to produce a final score that compares trackers.
We exploit the concepts of spatial overlap and cardinality of the METE measure~\cite{Nawaz2014} to formulate our measure.

\subsection{Intra-frame complexity}

\textit{The goal of the intra-frame complexity $E_{intra}$ is to quantify the effort the tracker makes to improve the detector's estimated bounding boxes for each frame}.
For each video sequence, let the intra-frame complexity be defined as
\begin{equation}
E_{intra} =  \frac{1}{K}\cdot\sum_{k=1}^{K} E_{intra}^k,
\label{eqn:Eintra}
\end{equation}
where $K$ is the total number of frames in the sequence and $E_{intra}^k \in \mathbb{R}_{[-1,1]}$ is the effort computed for each frame to measure the difference between the detector and tracker performances.
When $E_{intra}^k = 0$, the tracker maintains the same results as the detector. 
When $E_{intra}^k > 0$, the tracker improves the bounding boxes of the detector, for example new bounding boxes are created via temporal interpolation.
When $E_{intra}^k < 0$, the tracker worsens the bounding boxes of the detector, for example some bounding boxes are filtered out because deemed false positive detections.

The frame-level $E_{intra}^k$ is defined as
\begin{equation}
E_{intra}^k= \frac{1}{K}\cdot\sum_{k=1}^{K} (Q^k_t - Q^k_d),
\label{eqn:Eintra_k}
\end{equation}
where $Q^k_d$ is the performance achieved by the detector $d$ at frame $k$ that is defined as
\begin{equation}\label{eqn:Qd}
Q^k_d = I^k_d \cdot N^k_d = \left(1 - \frac{A^k_d}{L^k_d}\right) \cdot \left(1 - \frac{\left|V^k\right| - \left|U^k_d\right|}{max\left(\left|V^k\right|, \left|U^k_d\right|\right)}\right),
\end{equation}
where $I^k_d$ quantifies the similarity between the estimated and ground-truth bounding boxes.
$A^k_d$ is the total cost of the association between estimated and ground-truth bounding boxes.
This cost is computed by using the Hungarian algorithm based on the IOU criterion~\cite{kuhn1955hungarian}.

We normalize $A^k_d$ by the number of associated bounding boxes $L^k_d$. 
$\left|\cdot\right|$ is the cardinality of a set.
$N^k_d$ is the cardinality difference between the set of estimated bounding boxes $U_k^d$ and the set of ground-truth bounding boxes $V_k$. 

Analogously, $Q^k_t$ is the performance made by the tracker $t$ that we can obtain by substituting the set of estimated bounding boxes produced by $t$ at $k$, i.e.~$U_k^t$, to Eq.~\ref{eqn:Qd}. 

\begin{figure}[t]
\centering
\includegraphics[width=\columnwidth]{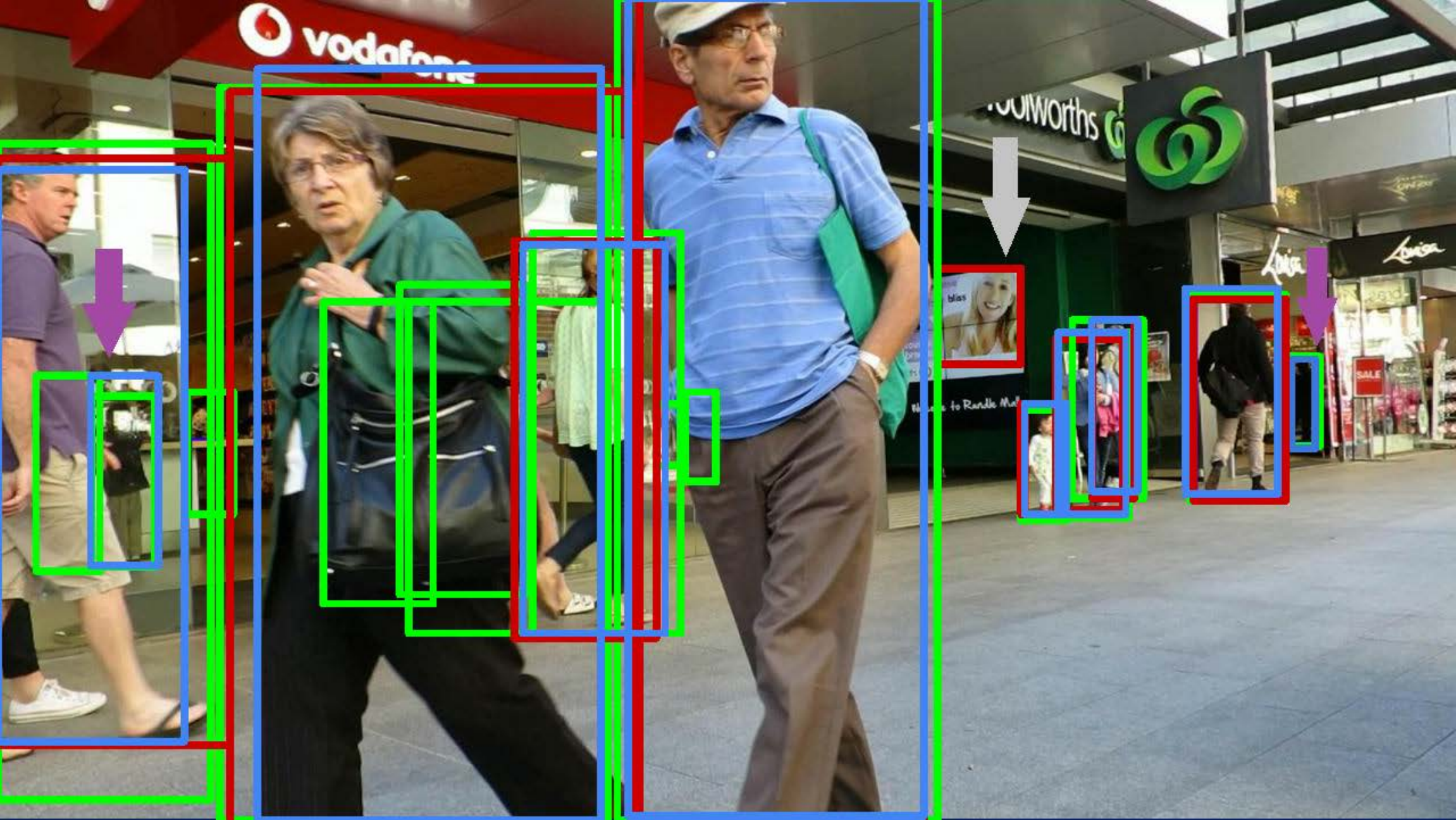}
\vspace{-5mm}
\caption{Detection and tracking results at $k$=180 of MOT17-09 sequence, by combining YoloV4 detector \cite{bochkovskiy2020yolov4} and DeepSORT tracker \cite{wojke2017simple}. 
The detector output, tracker output and ground-truth are represented by red, blue and green bounding boxes, respectively. 
Arrows show detector errors: false positive (gray) and false negatives (purple).}
\label{fig:intraframe-example-case1}
\end{figure}

Fig.~\ref{fig:intraframe-example-case1} shows an example of detection and tracking results at $k=180$.
The tracker corrects false positive detections, generated due to occlusions by pedestrians passing in front of the objects (one behind the man on the left of the image, near the counter in the shop; and the other is the bounding box furthest to the right). 
False Positives of the detector are removed by the tracker (gray arrow). 
Hence, the tracker is better than the detector, so $Q_t^{180}$- $Q_d^{180}$ is positive.

\subsection{Inter-frame complexity}
\textit{The goal of the inter-frame complexity $E_{inter}$ is to quantify the tracker's effort for correctly associating bounding boxes (i.e.~estimated detections or tracks) between frame pairs.} 
Let the inter-frame complexity be computed as
\begin{equation}\label{eqn:Einter}
E_{inter} = \frac{1}{K-1}\cdot\sum_{k=2}^{K} (Y^k + C^k \cdot IDSW_{score}^k),
\end{equation}
where $E_{inter}^k \in \mathbb{R}_{[-1,2]}$ and it considers two terms are considered for each frame: the improvement in the association tasks between consecutive frames ($Y^k\in \mathbb{R}_{[-1,1]}$) and 
the ID switches ($IDSW_{score}^k \in \mathbb{R}_{[0,1]}$) weighted by the mismatch in the number of objects considered ($C^k \in \mathbb{R}_{[0,1]}$).

Let the association improvement be defined as
\begin{equation}\label{eqn:yt}
Y^k = \left(1 - \frac{B^{k,k-1}_t}{L^{k,k-1}_t}\right) - \left(1 - \frac{B^{k,k-1}_d}{L^{k,k-1}_d}\right),
\end{equation}
where $B^{k,k-1}_d$ is the total cost for associating the bounding boxes between $k-1$ and $k$ frames for detection. Similarly $B^{k,k-1}_t$ is applied to tracking. If $B^{k,k-1}_t$ is lower than $B^{k,k-1}_d$, it means that the tracker states (i.e. bounding boxes) are better than the ones generated by the detector, in a inter-frame context.
Association costs are computed with the Hungarian Algorithm using the IOU criterion~\cite{kuhn1955hungarian} (as in Eq.~\ref{eqn:Qd}), and normalized by $L^{k,k-1}_d$ and $L^{k,k-1}_t$, which are the number of associated detection and tracking bounding boxes between consecutive frames, respectively.

Next, we consider the ID switches that may be produced by the tracker between frames $k-1$ and $k$.
Let the ID switch effort at $k$ be defined as
\begin{equation}\label{eqn:IDSW_score}
IDSW_{score}^k = \left(1 - \frac{IDSW^{k,k-1}}{L^{k,k-1}_t}\right),
\end{equation}
where $IDSW^{k,k-1}$ are the number of ID switches between frames $k-1$ and $k$, computed as defined in \cite{ristani2016performance}.

The $IDSW_{score}^k$ is weighted according to the number of tracked objects. Hence, we consider the cardinality of the IDs of the tracked objects as
\begin{equation}\label{eqn:cardinality_intra}
C^k =  \left(1 - \frac{\left|\mathcal{ID}^{k,k-1}\right|- L^{k,k-1}_t}{max(\left|\mathcal{ID}^{k,k-1}\right|, L^{k,k-1}_t)}\right),
\end{equation}
where $\mathcal{ID}^{k,k-1} = \mathcal{ID}^k \cap \mathcal{ID}^{k-1}$ is the set of unique ground truth IDs for frames $k$ and $k-1$. 
For example, if three ground-truth IDs (e.g.~$\mathcal{ID}^{k-1} = \{1,2,3\}$) exist at $k-1$ and three ground-truth IDs (e.g.~$\mathcal{ID}^k = \{3,4,5\}$) exist at $k$, then $|\mathcal{ID}^{k,k-1}|=5$.

\begin{figure}[t]
\centering
\includegraphics[width=\columnwidth]{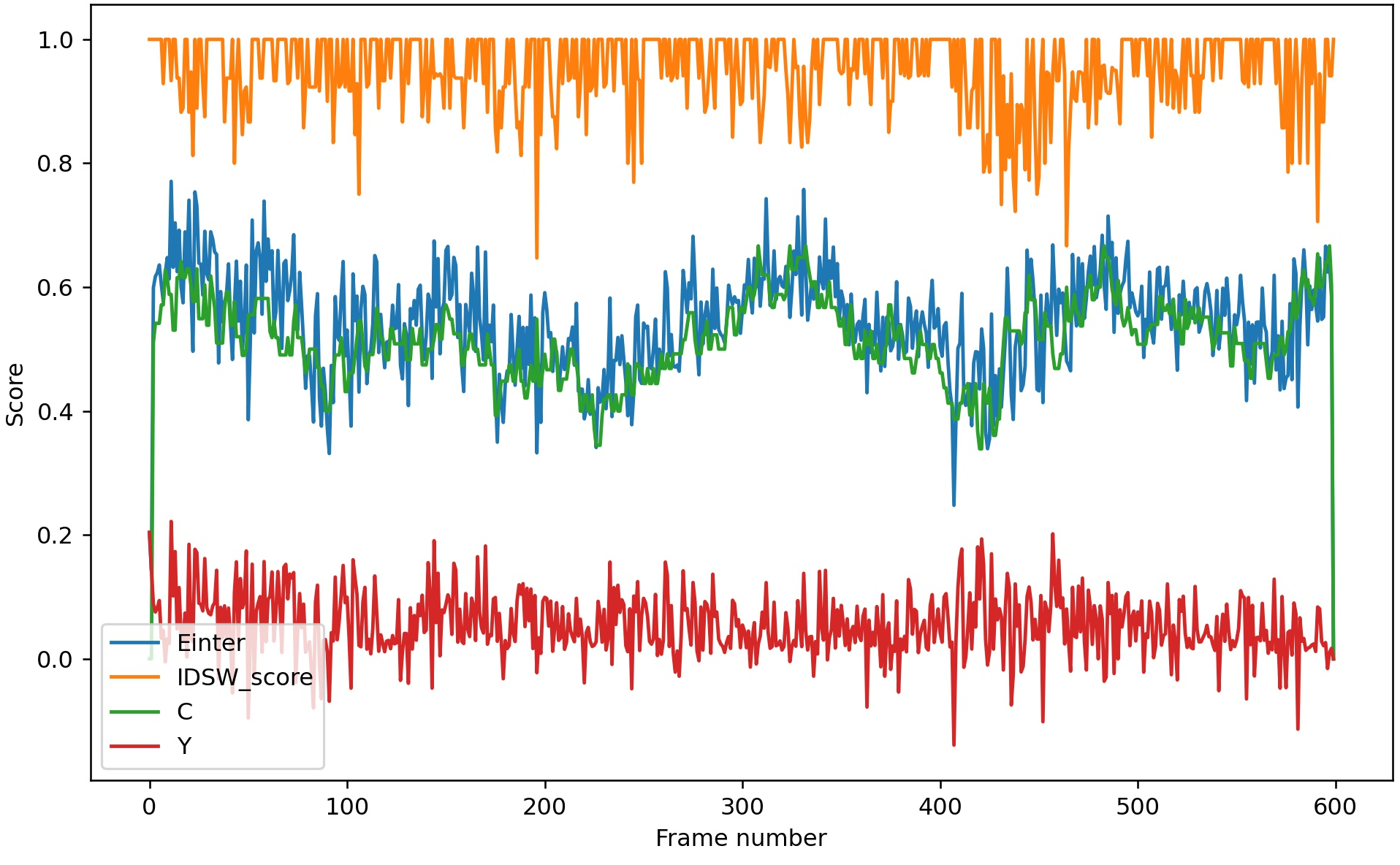}
\vspace{-5mm}
\caption{Results of \textit{inter-complexity measure} $E_{inter}$ and its components ($Y^k$, $C^k$ and $IDSW^k$) for the MOT17-02 sequence, by using the YoloV4 detector \cite{bochkovskiy2020yolov4} and the DeepSORT tracker \cite{wojke2017simple}.}
\label{fig:interframe-example-case1}
\end{figure}

Fig.~\ref{fig:interframe-example-case1} shows frame-level results for $E_{inter}$.
We can observe that few ID switches occur (i.e.~high values of the orange line).
The $Y^k$ component (red line) is positive on average, indicating that the tracker output makes easier the inter-frame association as compared to the detections. 
The tracker therefore corrects detector errors (i.e.~false positive and false negatives) in the inter-frame context.

\subsection{Final score}

The final effort applied by the tracker, is the combination of both, $E_{intra}$ and $E_{inter}$ complexity efforts. 
Both efforts are added since a unique performance score is desired for comparison.
Let the complexity score be defined as
\begin{equation}
\text{TEM} = \alpha \cdot E_{intra} + (1-\alpha) \cdot E_{inter},
\label{eqn:E}
\end{equation}
where $\alpha$ reweights the terms. To avoid any preference, we consider $\alpha=0.5$ in this work. 
Based on this choice of $\alpha$, TEM ranges between -1 (worst) and 1.5 (best).

\section{Experimental results}
\label{sec:evaluation}
We evaluate the proposed performance measures and compare them against conventional alternatives.

\subsection{Setup}

\subsubsection{Datasets}
We evaluate our evaluation measure by using two well-known datasets: MOT17~\cite{milan2016mot16} and MOT20~\cite{dendorfer2020mot20}. 
We select only the train sequences for both datasets because they have ground truth available. 
Our selection for MOT17 contains seven sequences with people walking that are captured by different camera angles, while MOT20 contains four sequences depicting crowded indoor and outdoor scenarios. 
We use 14,247 frames and 2,878 objects to track.

\subsubsection{Detectors and Trackers}
As we are interested in understanding how different detectors may affect performance measures for tracking, we generate different detection sets that produce different performances based on a realistic detector output.

Specifically, we employ the popular detector FasterRCNN~\cite{ren2015faster} and produce different sets of detections by changing two parameters, i.e.~\textit{box\_core\_thresh} and \textit{box\_nms\_thresh}.
The former is the minimum score allowed for a detection to be an object or not: the lower the threshold, the larger the number of false positive detections.
The latter is related to the non-maximum suppression step, which is in charge of removing duplicated (highly overlapping) bounding boxes of the same object.
A high value of \textit{box\_nms\_thresh} means that there will be more bounding boxes, so a larger number of false positives is expected.

Tab.~\ref{tab:faster-detections} shows the detection sets we generated by using this configurations, ranging from low performance (set \#5) to high performance (set \#6). 
We also consider the ground-truth detections as the set with perfect performance (set \#7). 
The set `fine-tune' corresponds to fine-tuning the detector using all the sequences. 
The set `public MOT' is the one provided by the organizers of the MOT challenge.

\begin{table}
\centering
\caption{Generated detection sets based on modifying two parameters of FasterRCNN~\cite{ren2015faster} (\textit{box\_core\_thresh} and \textit{box\_nms\_thresh}), for being used as input of trackers.}
\label{tab:faster-detections}
\vspace{-.2cm}
\resizebox{1\columnwidth}{!}{         
\begin{tabular}{lccc}
\toprule
FasterRCNN set & box\_score\_thresh & box\_nms\_thresh & mAP \\
\midrule
\#1 (default) & 0.05 & 0.5 & 0.55 \\
\#2 & 0 & 0 & 0.36 \\
\#3 & 0.5 & 0.5 & 0.49 \\
\#4 & 0.9 & 0.9 & 0.31 \\
\#5 & 0.99 & 0.5 & 0.18 \\
\#6 (fine-tune) & 0.05 & 0.5 & 0.72 \\
\#7 (ground-truth) & - & - & 1\\
\#8 public MOT & - & - & 0.49\\
\bottomrule
\end{tabular}
}
\vspace{-4mm}
\end{table}


As trackers, we employ four popular approaches: SORT \cite{bewley2016simple}, DeepSORT \cite{wojke2017simple}, UMA \cite{yin2020unified} and DAN \cite{sun2019deep}. 

By coupling detection sets (eight) and trackers (four), we obtain 32 combinations of detector-tracker results for the analysis presented in the following subsections.

\subsubsection{Performance evaluation measures}\label{sec:perf_eval}

We selected representative performance measures for comparison. For detection performance, we use mAP (mean Average Precision)~\cite{padilla2020survey}, Recall~\cite{nghiem2007etiseo}, Precision~\cite{nghiem2007etiseo}, TP (True Positives), FP (False Positives) and FN (False Negatives). For tracking performance, we use HOTA (Higher Order Tracking Accuracy) \cite{luiten2021hota} with an IOU score greater than 0 for association (i.e. HOTA(0)), MOTA (Multiple Object Tracking Accuracy) \cite{vcehovin2016visual}, MOTP (Multiple Object Tracking Precision) \cite{vcehovin2016visual}, IDF1 \cite{ristani2016performance}, ATA (Average Tracking Accuracy) \cite{manohar2006performance} and IDSW (Identity Switches) \cite{vcehovin2016visual}.

To compare the proposed performance measures with respect to related work, we use the Pearson product-moment correlation coefficient \cite{vcehovin2016visual}, which estimates the relation between two variables of different scale and ranges from -1 (inverse correlation) to +1 (direct correlation). To estimate the pair-wise correlation, we  get the 352 sequence tracking results (11 sequences analyzed with 32 detector-tracker combinations). Then, we select two performance measures and apply them to the 352 tracking results to obtain two performance scores for each sequence (by averaging the frame-level results). 
Note that performance measures for detection can also be applied to tracking results by removing the associated track identities from tracking results and ground-truth data (i.e.~only keeping the bounding boxes). 
Lastly, we use the Pearson coefficient to compare the two 352-dimensional vectors for the two selected measures.

\subsection{Analyzing existing detection and tracking performance evaluation measures}

\begin{figure}[t]
\begin{center}
  \includegraphics[width=1\linewidth]{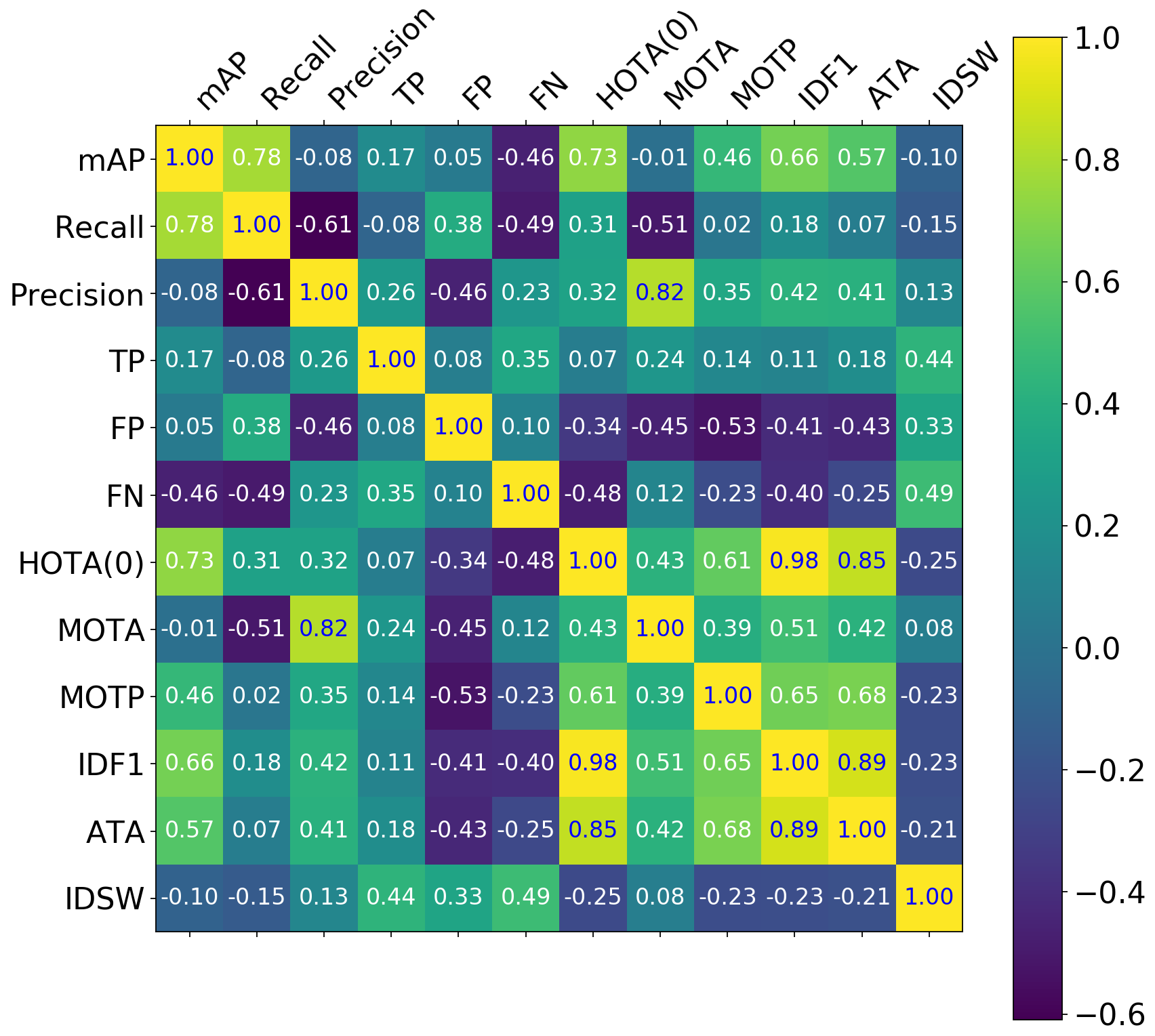}
\end{center}
  \vspace{-7mm}
  \caption{Correlation matrix for selected detection and tracking performance measures using 32 detector-tracker combinations over the MOT17 and MOT20 datasets.}
\label{fig:sota}
\end{figure}

Fig.~\ref{fig:sota} shows the result of the correlation analysis for the selected detection and tracking performance evaluation measures.
For the detection measures, we observe that \textit{mAP} has a high correlation with \textit{Recall}, and a low correlation with \textit{Precision}.
For the tracking measures, HOTA, IDF1 and ATA are highly correlated with each other. 
Our analysis also confirms the observation done in \cite{valmadre2021local}, which states that ATA is similar to IDF1.
MOTP results to be more correlated with IDF1 and ATA than with MOTA (MOTP measures the average localization accuracy over the TP set).

The correlation between detection and tracking measures shows the following. 
FP is inversely correlated with all tracking measures, but IDSW. 
FN is negatively correlated with HOTA, IDF1, MOTP and ATA. 
Unlike FN and IDSW, FN and MOTA are little correlated.

Moreover, we can observe that MOTA is highly correlated with Precision and negatively correlated with Recall, but not with mAP. 
This can be problematic as \textit{improving the precision of the detector would directly increase the MOTA score}, which is a behavior that was also observed in \cite{luiten2021hota}.
Also MOTP, IDF1 and ATA have some degree of correlation with Precision.
HOTA is highly correlated with mAP. 
In summary, we empirically observed that \textit{the detection performance can somewhat affect the evaluation of the tracking performance, thus making the comparison among trackers difficult to be carried out accurately}.

\subsection{Results for intra-frame complexity}

\begin{figure}[t]
\begin{center}
  \includegraphics[width=1\linewidth]{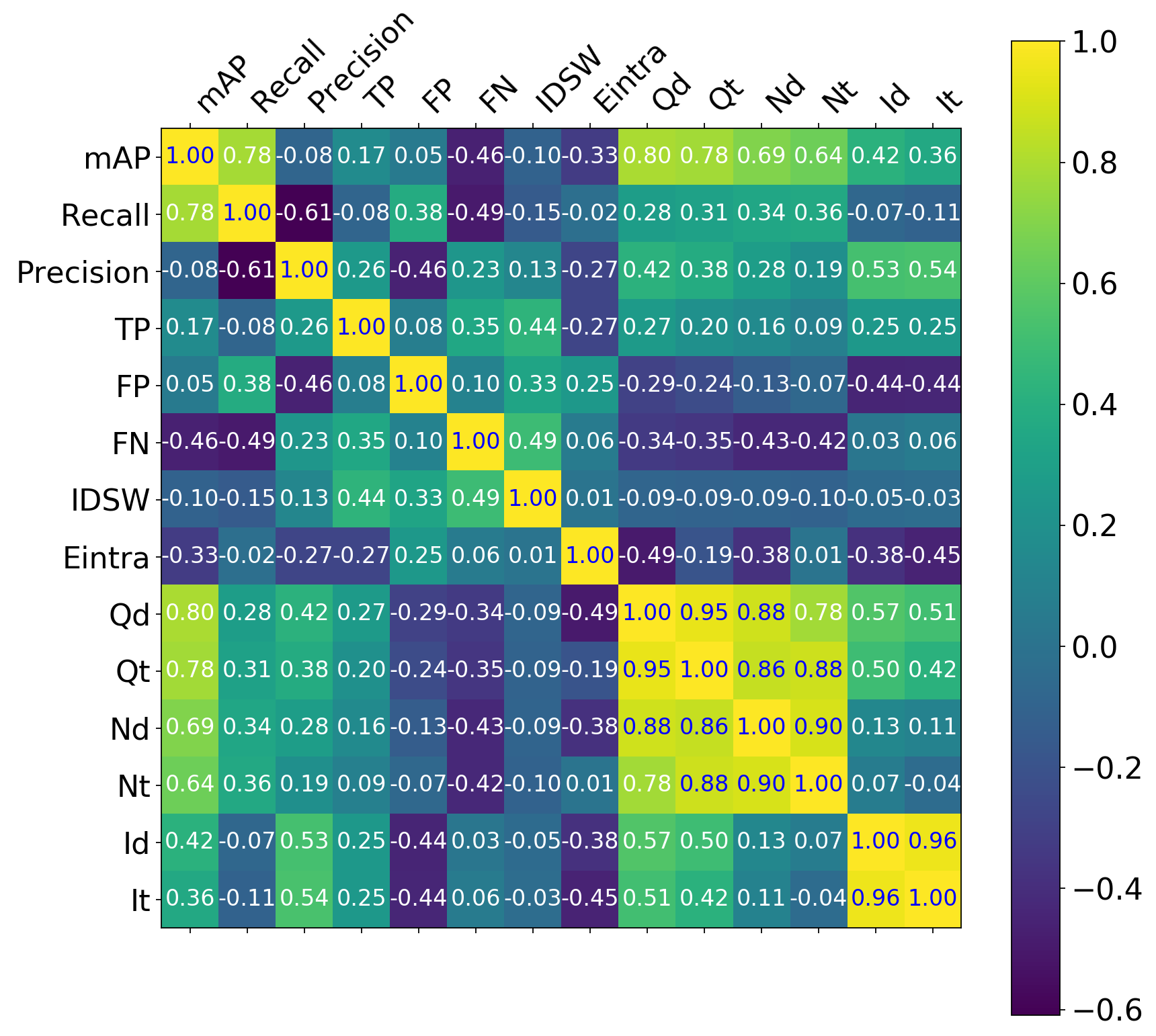}
\end{center}
  \vspace{-7mm}
  \caption{Correlation matrix for the proposed \textit{intra-complexity measure} $E_{intra}$ and its components ($Q_d$, $Q_t$, $N_d$, $N_t$, $I_d$ and $I_t$), using 32 detector-tracker combinations over the MOT17 and MOT20 datasets.}
\label{fig:Eintra}
\end{figure}

Fig.~\ref{fig:Eintra} shows the correlation matrix for the proposed intra-frame complexity measure $E_{intra}$ and its components. 
Only detection metrics are included because only the matching between estimated and ground-truth bounding boxes is measured. 
We can observe two highly correlated clusters (\{$Q_d$, $Q_t$, $N_d$ and $N_t$\} and \{$I_d$ and $I_t$\}). 
The correlation between $E_{intra}$ and $Q_d$ is negative, while the correlation between $E_{intra}$ and $Q_t$ is closer to zero.
This means that the tracker improves the detector’s predictions on average.
$Q_d$ and $Q_t$ are strongly correlated with mAP, which makes them very prone to changes in the detector's performance. 
However, we can observe that their difference (i.e.~$E_{intra}$) exhibits a lower correlation (in absolute value) with mAP, Precision and Recall. Therefore, the \textit{low correlation of $E_{intra}$ with the detection measures shows its effectiveness in evaluating the work done by the tracker, removing the dependency with detector's performance}.

\subsection{Results for inter-frame complexity}

\begin{figure}[t]
\begin{center}
\includegraphics[width=1\linewidth]{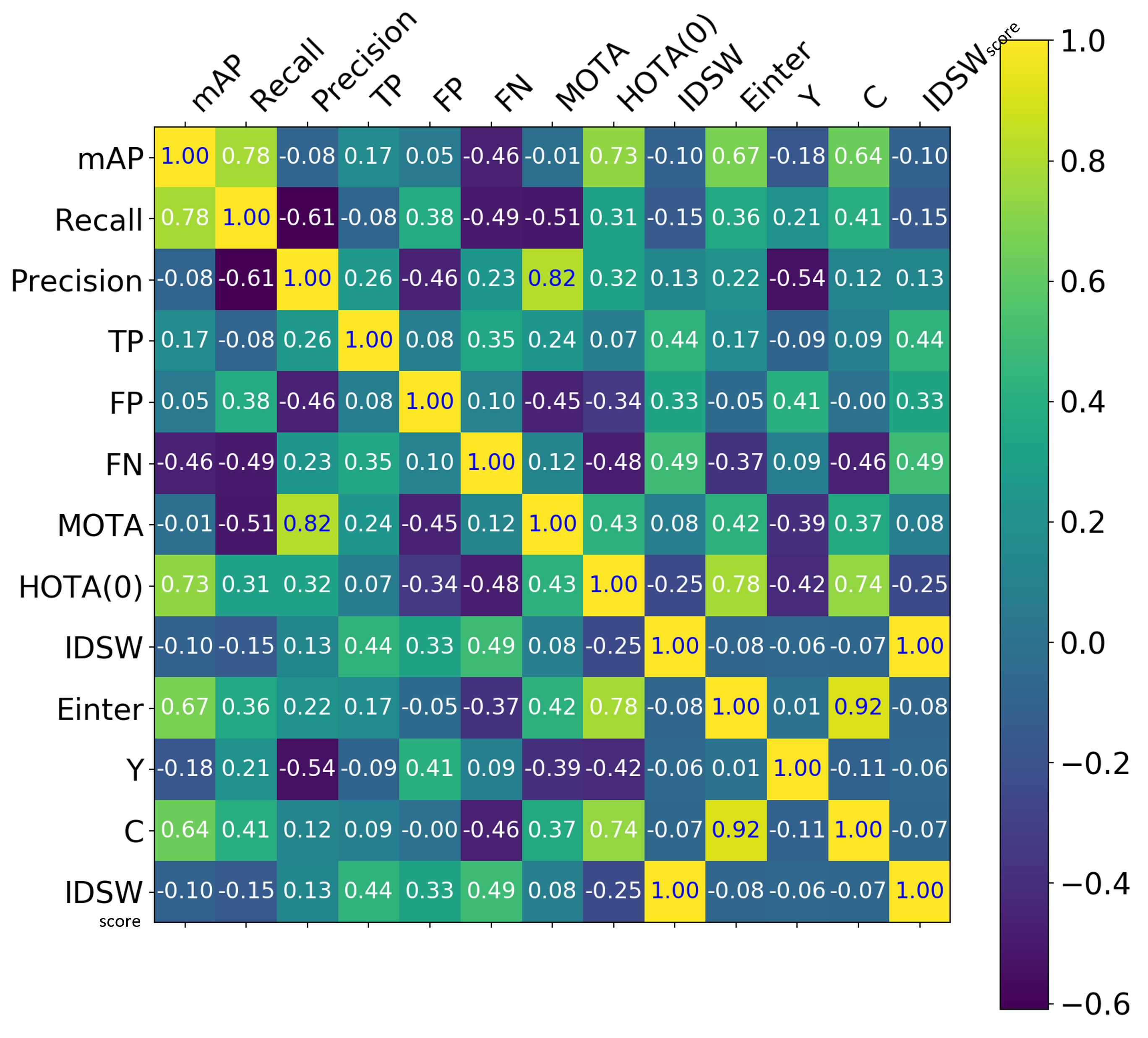}
\end{center}
\vspace{-7mm}
\caption{Correlation matrix for the proposed \textit{inter-complexity measure} $E_{inter}$ and its components ($Y$, $C$ and $IDSW$), using 32 detector-tracker combinations over the MOT17 and MOT20 datasets.}
\label{fig:Einter}
\end{figure}

Fig.~\ref{fig:Einter} shows the correlation matrix for the inter-frame complexity measure $E_{inter}$ and its components. 
We can observe that $E_{inter}$ is highly correlated with $C$, which quantifies the cardinality (Eq.~\ref{eqn:cardinality_intra}), while it is little correlated with $Y$ and $IDSW$, suggesting that $C$ dominates in $E_{inter}$.

For the detection measures, $E_{inter}$ is correlated to mAP, and slightly less correlated with Recall and Precision.
However, the correlation with Recall and Precision is similar, which is 0.36 and 0.22, respectively.
This is another outcome we are looking for because \textit{a non-equal weight between Recall and Precision may be problematic}. 
Although also HOTA has similar correlation with Precision and Recall, the Precision correlation values are higher than that of $E_{inter}$, thus implying a stronger dependency on detector's performance.
Although $Y$ is negatively correlated to Precision, it does not significantly affect $E_{inter}$ due to its lower values as compared to the other components (see Fig. \ref{fig:interframe-example-case1}).

For the tracking measures, we observe that $E_{inter}$ has some degree of correlation with HOTA and, to some extent, with MOTA too.
This also occurs for the $C$ component. 
In contrast, $Y$ is inversely correlated with HOTA and MOTA.

In summary, \textit{the effort calculated in the context of inter-frame complexity $E_{inter}$ is valid and meets our expectations as the correlation between the $E_{inter}$ and Precision/Recall is balanced.}
However, the correlation of $E_{inter}$ with mAP is high (albeit lower than HOTA) and may make difficult to evaluate detector-tracker combinations. Similar conclusions can be obtained for the correlation between Recall and $E_{inter}$, with respect to Recall and MOTA.

\subsection{Results for final complexity}

TEM is calculated by combining intra-frame and inter-frame efforts with $\alpha = 0.5$ (Eq.~\ref{eqn:E}).

\begin{figure}[t]
\begin{center}
\includegraphics[width=1\linewidth]{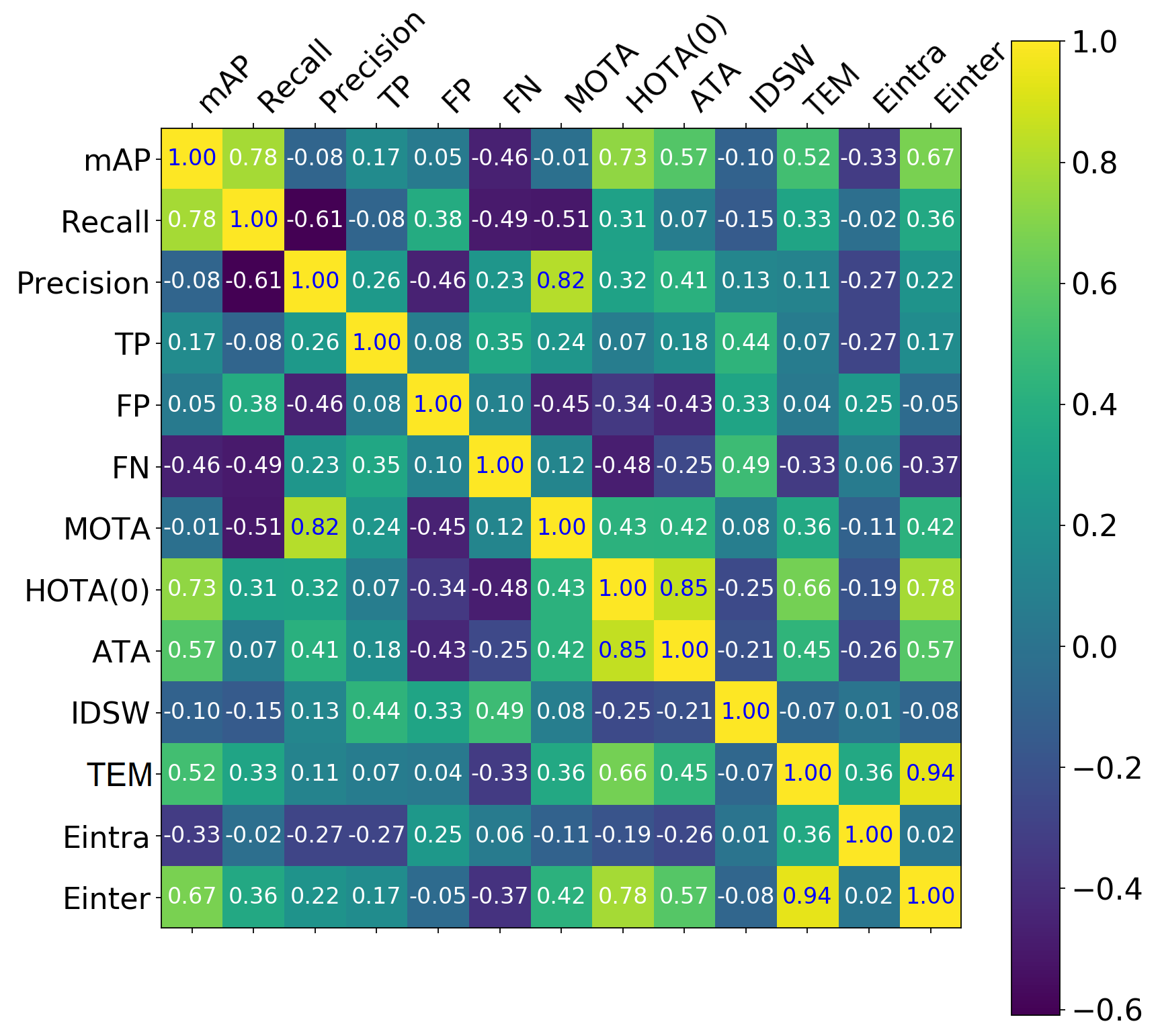}
\end{center}
\vspace{-7mm}
\caption{Correlation matrix for the proposed \textit{complexity measure} TEM that combines the $E_{intra}$ and $E_{inter}$, using the 32 detector-tracker combinations over the MOT17 and MOT20 datasets.}
\label{fig:E}
\end{figure}

Fig.~\ref{fig:E} shows the correlation matrix comparing detection and tracking measures with the proposed measure.
On the one hand, we can observe that TEM and $E_{inter}$ are highly correlated between them. 
On the other hand, $E_{intra}$ is less correlated than $E_{inter}$, with respect to TEM.
This is because the values of $E_{intra}$ are smaller than the $E_{inter}$, so it weights less in the final effort (TEM).
As compared to other tracking measures, we can observe that TEM is somehow correlated with HOTA, MOTA and ATA, implying that the new measure can effectively evaluate tracking performance.
However, we successfully reduced the dependency on detector's performance (mAP, Precision and Recall) as it results to be lower for the proposed measure TEM as compared to both HOTA and MOTA.

Tab.~\ref{tab:metrics_sort_table_effort} compares detection measures (mAP, Recall, Precision) and tracking measures (HOTA, ATA, TEM) in the case of six sequences extracted from MOT17 and MOT20 datasets for for two detection sets of \textit{Faster R-CNN} \cite{ren2015faster} and the SORT tracker \cite{bewley2016simple}.
HOTA's main trend is related to Recall/mAP (i.e.~the higher Recall/mAP, the lower HOTA). However, some sequences like MOT17-
02 and MOT17-11 show an opposite outcome with respect to mAP.
We can observe a non-consistent behavior of HOTA with respect to Precision. Albeit most of the sequences are directly correlated, some sequences like MOT20-01 and MOT20-02 show an opposite relationship between Prediction and HOTA. 
The proposed TEM measure shows a consistent trend for all sequences, being positively correlated with mAP. Moreover, the improvement in precision does not seem to affect the measure as much as it does in HOTA, giving higher scores when the detector improves mAP. 
Moreover, detection results for each sequence exhibit similar mAP for sets \#1 and \#3. However precision is always higher for set \#3, which means less FPs. A lower number of FPs implies that the effort of the tracker should be less as the mAP is similar in the two cases. Unlike HOTA/ATA, the proposed TEM is able to capture such effort, giving lower scores when Precision is higher.

\begin{table}
\tabcolsep 4pt
\caption{Comparison of the performance of tracking measures, with the proposed measure, when using the SORT tracker \cite{bewley2016simple} and two detection sets for \textit{Faster R-CNN} \cite{ren2015faster} from Table \ref{tab:faster-detections}.}
\label{tab:metrics_sort_table_effort}
\vspace{-.2cm}
\resizebox{\columnwidth}{!}{
\begin{tabular}{lccccccc}
\toprule
Sequence  & Detector & mAP & Recall & Precision & HOTA(0) & ATA & TEM \\
\midrule
\multirow{2}{*}{MOT17-02} & FasterRCNN\#1 & 0.41 & 0.41 & 0.59 & 0.31 & 0.01 & 0.78 \\ 
                      & FasterRCNN\#3 & 0.34 & 0.24 & 0.90 & 0.35 & 0.06 & 0.59 \\ 
\midrule
\multirow{2}{*}{MOT17-05} & FasterRCNN\#1 & 0.64 & 0.68 & 0.45 & 0.46 & 0.06 & 1.02 \\ 
                      & FasterRCNN\#3 & 0.60 & 0.48 & 0.92 & 0.58 & 0.20 & 0.84 \\
\midrule
\multirow{2}{*}{MOT17-09} & FasterRCNN\#1 & 0.67 & 0.67 & 0.58 & 0.47 & 0.03 & 0.97 \\ 
                     & FasterRCNN\#3 & 0.64 & 0.50 & 0.97 & 0.59 & 0.17 & 0.90 \\
\midrule
\multirow{2}{*}{MOT17-11} & FasterRCNN\#1 & 0.67 & 0.70 & 0.43 & 0.48 & 0.03 & 0.97 \\ 
                     & FasterRCNN\#3 & 0.64 & 0.54 & 0.95 & 0.60 & 0.13 & 0.94 \\
\midrule
\multirow{2}{*}{MOT20-01} & FasterRCNN\#1 & 0.57 & 0.47 & 0.88 & 0.49 & 0.05 & 0.87 \\ 
                      & FasterRCNN\#3 & 0.49 & 0.29 & 0.97 & 0.46 & 0.12 & 0.67 \\
\midrule
\multirow{2}{*}{MOT20-02} & FasterRCNN\#1 & 0.54 & 0.41 & 0.90 & 0.41 & 0.04 & 0.85 \\ 
                      & FasterRCNN\#3 & 0.45 & 0.26 & 0.97 & 0.38 & 0.08 & 0.67 \\
\bottomrule
\end{tabular}
}
\end{table}

\section{Conclusions}\label{sec:conclusions}

We explored the evaluation of tracking algorithms employing different detectors by proposing a new performance evaluation measure that disentangles the performance of the detector from that of the tracker. 
Our new performance evaluation measure accounts for the effort done by the tracker, given different sets of detections with different performances.
This measure is based on spatial overlap and cardinality estimations, with the objective of analyzing the tracker effort at two levels: intra-complexity (frame) and inter-complexity (sequence or consecutive frames), by comparing the tracking output with the ground truth.
The experimental results identified correlations among existing detection and tracking measures, exhibiting a strong dependency of tracking performance measures with respect to some performance indicators of the detection set. The proposed measures decrease such dependency on detection, making possible to evaluate trackers with different detection sets. Moreover, non-uniform behaviors were observed for well-known measures, unlike the proposed measures which show a consistent behavior. The proposed measures do not pretend to substitute existing measures, but to provide an alternative viewpoint for evaluating different combinations of detectors and trackers. 
As future work, we will focus on extending the experiments to other detectors and trackers, improving the inter-complexity to reduce its detection dependency and also on designing proper combinations of the proposed performance measures.

\noindent\textbf{Limitation.}
One limitation of TEM is related to the terms of Eq.~\ref{eqn:Einter} that are not equally balanced and this can be observed in the experiment of Fig.~\ref{fig:interframe-example-case1}.
$Y^k$ is significantly lower than $IDSW^k$ and $C^k$.
We mitigated this problem by formulating $E_{inter}$ as the sum between $Y$ and the product between $IDSW$ and $C$.
We deem that a more effective way of combining this terms can be formulated.


{\small
\bibliographystyle{ieee}
\bibliography{egbib}

\begin{thebibliography}{10}\itemsep=-1pt

\bibitem{bernardin2008evaluating}
K.~Bernardin and R.~Stiefelhagen.
\newblock Evaluating multiple object tracking performance: the clear mot
  metrics.
\newblock {\em EURASIP Journal on Image and Video Processing}, 2008:1--10,
  2008.

\bibitem{bewley2016simple}
A.~Bewley, Z.~Ge, L.~Ott, F.~Ramos, and B.~Upcroft.
\newblock Simple online and realtime tracking.
\newblock In {\em Proc. of IEEE Int. Conf. on Image Processing}, pages
  3464--3468, 2016.

\bibitem{bochkovskiy2020yolov4}
A.~Bochkovskiy, C.-Y. Wang, and H.-Y.~M. Liao.
\newblock Yolov4: Optimal speed and accuracy of object detection.
\newblock {\em arXiv preprint arXiv:2004.10934}, 2020.

\bibitem{Carr2016}
P.~Carr and R.~Collins.
\newblock Assessing tracking performance in complex scenarios using mean time
  between failures.
\newblock In {\em Proc. of IEEE Winter Conf. on Applications of Computer
  Vision}, pages 1--6, 2016.

\bibitem{carvalho2012filling}
P.~Carvalho, J.~S. Cardoso, and L.~Corte-Real.
\newblock Filling the gap in quality assessment of video object tracking.
\newblock {\em Image and Vision Computing}, 30(9):630--640, 2012.

\bibitem{vcehovin2016visual}
L.~{\v{C}}ehovin, A.~Leonardis, and M.~Kristan.
\newblock Visual object tracking performance measures revisited.
\newblock {\em IEEE Trans. on Image Processing}, 25(3):1261--1274, 2016.

\bibitem{chen2022survey}
F.~Chen, X.~Wang, Y.~Zhao, S.~Lv, and X.~Niu.
\newblock Visual object tracking: A survey.
\newblock {\em Computer Vision and Image Understanding}, 222:103508, 2022.

\bibitem{conte2010performance}
D.~Conte, P.~Foggia, G.~Percannella, and M.~Vento.
\newblock Performance evaluation of a people tracking system on {PETS}2009
  database.
\newblock In {\em Proc. of IEEE Int. Conf. on Advanced Video and Signal-based
  Surveillance}, pages 119--126, 2010.

\bibitem{dendorfer2021motchallenge}
P.~Dendorfer, A.~Osep, A.~Milan, K.~Schindler, D.~Cremers, I.~Reid, S.~Roth,
  and L.~Leal-Taix{\'e}.
\newblock {MOTChallenge: A Benchmark for Single-Camera Multiple Target
  Tracking}.
\newblock {\em Int. Journal of Computer Vision}, 129(4):845--881, 2021.

\bibitem{dendorfer2020mot20}
P.~Dendorfer, H.~Rezatofighi, A.~Milan, J.~Shi, D.~Cremers, I.~Reid, S.~Roth,
  K.~Schindler, and L.~Leal-Taix{\'e}.
\newblock Mot20: A benchmark for multi object tracking in crowded scenes.
\newblock {\em arXiv preprint arXiv:2003.09003}, 2020.

\bibitem{denman2009dynamic}
S.~Denman, C.~Fookes, S.~Sridharan, and R.~Lakemond.
\newblock Dynamic performance measures for object tracking systems.
\newblock In {\em Proc. of IEEE Int. Conf. on Advanced Video and Signal-based
  Surveillance}, pages 541--546, 2009.

\bibitem{Everingham2010}
M.~Everingham, L.~Van~Gool, C.~Williams, and A.~Winn, J. abd~Zisserman.
\newblock The pascal visual object classes (voc) challenge.
\newblock {\em International Journal on Computer Vision}, (88):303--338, 2010.

\bibitem{fang2017performance}
Y.~Fang, Y.~Yuan, L.~Li, J.~Wu, W.~Lin, and Z.~Li.
\newblock Performance evaluation of visual tracking algorithms on video
  sequences with quality degradation.
\newblock {\em IEEE Access}, 5:2430--2441, 2017.

\bibitem{garcia2021time}
{\'A}.~Garc{\'\i}a-Fern{\'a}ndez, A.~Rahmathullah, and L.~Svensson.
\newblock A time-weighted metric for sets of trajectories to assess
  multi-object tracking algorithms.
\newblock In {\em Proc. of IEEE Int. Conf. on Information Fusion}, pages 1--8,
  2021.

\bibitem{kasturi2008framework}
R.~Kasturi, D.~Goldgof, P.~Soundararajan, V.~Manohar, J.~Garofolo, R.~Bowers,
  M.~Boonstra, V.~Korzhova, and J.~Zhang.
\newblock Framework for performance evaluation of face, text, and vehicle
  detection and tracking in video: Data, metrics, and protocol.
\newblock {\em IEEE Trans. on Pattern Analysis and Machine Intelligence},
  31(2):319--336, 2008.

\bibitem{kuhn1955hungarian}
H.~Kuhn.
\newblock The hungarian method for the assignment problem.
\newblock {\em Naval research logistics quarterly}, 2(1-2):83--97, 1955.

\bibitem{luiten2021hota}
J.~Luiten, A.~Osep, P.~Dendorfer, P.~Torr, A.~Geiger, L.~Leal-Taix{\'e}, and
  B.~Leibe.
\newblock Hota: A higher order metric for evaluating multi-object tracking.
\newblock {\em Int. Journal of Computer Vision}, 129(2):548--578, 2021.

\bibitem{manohar2006performance}
V.~Manohar, P.~Soundararajan, H.~Raju, D.~Goldgof, R.~Kasturi, and J.~Garofolo.
\newblock Performance evaluation of object detection and tracking in video.
\newblock In {\em Proc. of Asian Conf. on Computer Vision}, pages 151--161,
  2006.

\bibitem{marcenaro2012performance}
L.~Marcenaro, P.~Morerio, and C.~S. Regazzoni.
\newblock Performance evaluation of multi-camera visual tracking.
\newblock In {\em Proc. of IEEE Int. Conf. on Advanced Video and Signal-based
  Surveillance}, pages 464--469, 2012.

\bibitem{marvasti2021deep}
S.~Marvasti-Zadeh, L.~Cheng, H.~Ghanei-Yakhdan, and S.~Kasaei.
\newblock Deep learning for visual tracking: A comprehensive survey.
\newblock {\em IEEE Trans. Intelligent Transportation Systems},
  23(5):3943–3968, 2021.

\bibitem{milan2016mot16}
A.~Milan, L.~Leal-Taix{\'e}, I.~Reid, S.~Roth, and K.~Schindler.
\newblock Mot16: A benchmark for multi-object tracking.
\newblock {\em arXiv preprint arXiv:1603.00831}, 2016.

\bibitem{Nawaz2014}
T.~Nawaz, F.~Poiesi, and A.~Cavallaro.
\newblock Measures of effective video tracking.
\newblock {\em IEEE Trans. on Image Processing}, 23(1):1--13, Jan. 2014.

\bibitem{nghiem2007etiseo}
A.-T. Nghiem, F.~Bremond, M.~Thonnat, and V.~Valentin.
\newblock Etiseo, performance evaluation for video surveillance systems.
\newblock In {\em Proc. of IEEE Conf. on Advanced Video and Signal-based
  Surveillance}, pages 476--481, 2007.

\bibitem{padilla2020survey}
R.~Padilla, S.~Netto, and E.~Da~Silva.
\newblock A survey on performance metrics for object-detection algorithms.
\newblock In {\em Proc. of Int. Conf. on Systems, Signals and Image
  Processing}, pages 237--242, 2020.

\bibitem{pinto2021uncertainty}
J.~Pinto, Y.~Xia, L.~Svensson, and H.~Wymeersch.
\newblock An uncertainty-aware performance measure for multi-object tracking.
\newblock {\em IEEE Signal Processing Letters}, 28:1689--1693, 2021.

\bibitem{ren2015faster}
S.~Ren, K.~He, R.~Girshick, and J.~Sun.
\newblock Faster r-cnn: Towards real-time object detection with region proposal
  networks.
\newblock {\em Proc. of Advances in Neural Information Processing Systems}, 28,
  2015.

\bibitem{ristani2016performance}
E.~Ristani, F.~Solera, R.~Zou, R.~Cucchiara, and C.~Tomasi.
\newblock Performance measures and a data set for multi-target, multi-camera
  tracking.
\newblock In {\em Proc. of European Conf. on Computer Vision}, pages 17--35,
  2016.

\bibitem{sanmiguel2012adaptive}
J.~C. SanMiguel, A.~Cavallaro, and J.~M. Mart{\'\i}nez.
\newblock Adaptive online performance evaluation of video trackers.
\newblock {\em IEEE Trans. on Image Processing}, 21(5):2812--2823, 2012.

\bibitem{sun2019deep}
S.~Sun, N.~Akhtar, H.~Song, A.~Mian, and M.~Shah.
\newblock Deep affinity network for multiple object tracking.
\newblock {\em IEEE Trans. Pattern Analysis and Machine Intelligence},
  43(1):104--119, 2019.

\bibitem{valmadre2021local}
J.~Valmadre, A.~Bewley, J.~Huang, C.~Sun, C.~Sminchisescu, and C.~Schmid.
\newblock Local metrics for multi-object tracking.
\newblock {\em arXiv preprint arXiv:2104.02631}, 2021.

\bibitem{wojke2017simple}
N.~Wojke, A.~Bewley, and D.~Paulus.
\newblock Simple online and realtime tracking with a deep association metric.
\newblock In {\em Proc. of IEEE Int. Conf. on Image Processing}, pages
  3645--3649, 2017.

\bibitem{yin2020unified}
J.~Yin, W.~Wang, Q.~Meng, R.~Yang, and J.~Shen.
\newblock A unified object motion and affinity model for online multi-object
  tracking.
\newblock In {\em Proc. of IEEE/CVF Conf. on Computer Vision and Pattern
  Recognition}, pages 6768--6777, 2020.

\bibitem{zaidi2022survey}
S.~Zaidi, M.~Ansari, A.~Aslam, N.~Kanwal, M.~Asghar, and B.~Lee.
\newblock A survey of modern deep learning based object detection models.
\newblock {\em Digital Signal Processing}, page 103514, 2022.

\end{thebibliography}
}


\end{document}